\documentclass{article}

\usepackage{microtype}
\usepackage{graphicx}
\usepackage{subfigure}
\usepackage{booktabs} 
\usepackage[]{algorithm2e}
\usepackage{amssymb,mathtools}
\usepackage{mathptmx}
\usepackage{graphicx}
\usepackage{times}
\usepackage{enumitem}

\usepackage{hyperref}

\usepackage[accepted]{icml2020}

\icmltitlerunning{Visualizing Classification Structure of Large-Scale Classifiers}

\begin{document}

\twocolumn[
\icmltitle{Visualizing Classification Structure of Large-Scale Classifiers}



\icmlsetsymbol{equal}{*}

\begin{icmlauthorlist}
\icmlauthor{Bilal Alsallakh}{to,goo}
\icmlauthor{Zhixin Yan}{goo}
\icmlauthor{Shabnam Ghaffarzadegan}{goo}
\icmlauthor{Zeng Dai}{goo}
\icmlauthor{Liu Ren}{goo}
\end{icmlauthorlist}

\icmlaffiliation{goo}{Bosch Research North America}
\icmlaffiliation{to}{Facebook, Menlo Park, USA}

\icmlcorrespondingauthor{Bilal Alsallakh}{bilalsal@fb.com}

\icmlkeywords{Machine Learning, ICML}

\vskip 0.3in
]

\printAffiliationsAndNotice{}  

\begin{abstract}

We propose a measure to compute class similarity in large-scale classification based on prediction scores.
Such measure has not been formally proposed in the literature.
We show how visualizing the class similarity matrix can reveal hierarchical structures and relationships that govern the classes. Through examples with various classifiers, we demonstrate how such structures can help in analyzing the classification behavior and in inferring potential corner cases.
The source code for one example is available as a notebook at \href{https://github.com/bilalsal/blocks}{https://github.com/bilalsal/blocks.}
\end{abstract}

\section{Introduction}

\label{sec:intro}

Classes that are highly similar are harder to separate from each other than from other classes.
A variety of methods leverage this insight to improve multi-class classifiers~\cite{amit2007uncovering, deng2014large, murdock2016blockout}.
The curators of ImageNet noted that visual object categories inherently follow a hierarchical similarity structure that is reflected in the confusion matrix of various ImageNet classifiers~\cite{deng2010does}.
Recent work demonstrates how this structure is further reflected in the features learned at successive layers in deep neural networks \cite{alsallakh2017convolutional}.

Inspired by the above-mentioned observations, our goal is to provide generic means to analyze and visualize class similarity structure in large-scale classification.
Such analysis helps understand the features learned by a classifier to discriminate between classes.
Confusion matrices fall short of enabling such analysis in a generic way.
Our contributions include (1)
  a novel class similarity measure based on prediction scores, described in Section~\ref{sec:rethinking}, and
(2) means to visualize the class similarity matrix with examples of structures this matrix can reveal in three large-scale datasets, described in Section~\ref{sec:vis}.


\section{Rethinking Class Similarity}
\label{sec:rethinking}

Both pieces of work mentioned in Section~\ref{sec:intro} rely on confusion matrices to analyze classification structure \cite{alsallakh2017convolutional, deng2010does}.
When ordered according to the ImageNet synset hierarchy, this matrix captures the majority of confusions in few diagonal blocks that correspond to coarse similarity groups.
Each of these blocks, in turn, can exhibit a nested block pattern that corresponds to narrower groups in the hierarchy.
As we illustrate  in Section~\ref{sec:vis},  confusion matrices might fail to exhibit such pattern that reflects similarity structures due to the following reasons:
\begin{itemize}
\item {\textbf{Sparsity:}}
Large confusion matrices usually contain more cells than labeled data, leaving the majority of cells empty, even with a relatively high error rate.
\item {\textbf{Class imbalance:}}
When reordering an imbalanced confusion matrix, the computed order is determined mainly by over-represented classes.
This limits the possibilities to explore similarities involving
under-represented classes, even if they constitute the majority.
\item {\textbf{Multi-label classification:}}
Confusion matrices are ill-defined in such problems.
A multi-label classifier makes an error either by missing one label from the multi-labeled ground truth or by predicting a superfluous label.
In both cases, the error cannot be attributed to a pair of classes as in a confusion matrix.

\end{itemize}
These characteristics are prevalent in large-scale classification as we exemplify with three popular datasets. The appendix provides two further examples.

The limited applicability of confusion matrices for similarity analysis is due to the reliance on pair-wise class confusions as a proxy of class similarity.
If the classifier makes no errors, the matrix will be diagonal and hence will fail to show underlying class similarities.
Moreover, when an error occurs, only the top-1 prediction of the classifier record the information.
This discards the remaining top-k guesses of the classifier whose scores contain rich information about the result.
Hinton et al. leverage these scores to distill essential knowledge about a classifier~\cite{hinton2015distilling}, noting that they carry rich information about similarity structure in the data.
We use these scores to analyze and visualize class similarities.

\subsection*{Similarity based on Prediction Scores}

\label{sec:simPred}
Given a set of $m$ classes $C = \{c_j : 1 \le j \le m\}$,
we treat the scores $p_{i,j}$ computed by the classifier for each class $c_j$ as a random variable $P_j$ over the set of samples $X = \{x_i: 1 \le i \le n\}$.
This enables us to formulate class similarity as correlation between these variables.
Several measures of dependence between two variables have been defined.
{Pearson's correlation coefficient} measures the linear relationship between two random variables as follows:
\begin{equation}
	\rho_{j_1, j_2} = \frac{E(P_{j_1} \cdot P_{j_2})- E(P_{j_1}) \cdot E(P_{j_2})}{\sqrt{{E(P_{j_1}}^2) - E(P_{j_1})^2} \cdot \sqrt{{E(P_{j_2}}^2) - E(P_{j_2})^2}}
\label{eq:sim-Pearson}
\end{equation}
where $E$ denotes the expected value and is computed as the mean over all $x_i \in X$.
The expected value $E(P_{j_1} \cdot P_{j_2})$ is high when $c_{j_1}$ and $c_{j_2}$ frequently appear \emph{together} among the top guesses, which supports their overall similarity.
It is worth mentioning that we do not use the computed correlations to perform statistical significance test.
Instead, we use these correlations to define a class similarity matrix $M$ that has several  advantages over a confusion matrix:

\begin{itemize}
\item
A prediction does not need to be erroneous for a sample to contribute to the similarity computation, alleviating the sparsity problem of the computed matrix.
\item
No ground-truth labels are required, as evident in Eq.~\ref{eq:sim-Pearson}.
This enables computing class similarity over unlabeled datasets as long as a prototypical classifier is available.
\item
The coefficient can be naturally applied to multi-labeled datasets because they do not rely on comparing ground truth with predictions.
\item
The coefficient is symmetric with respect to the classes and is normalized with respect to their frequencies.
Both properties are crucial to visualize possible block patterns in $M$ and to handle class imbalance.
\end{itemize}

\subsection*{Statistical Analysis of the Similarity Measure}

\label{sec:statAnalysis}

It is important to pay attention to the distribution of the similarity values.
As we demonstrate in Section~\ref{sec:vis}, a nearly-normal distribution of the values in $M$ enables both small and large similarity groups to emerge when visualizing M.
For illustration, we examine this condition on the \texttt{Places365} dataset~\cite{zhou2017places}.
This dataset contains 365 classes that represent scenes and places.
We use a pre-trained AlexNet classifier provided by the dataset curators to compute prediction scores and model activations for all 36500 samples in the validation set.

\begin{figure}[!ht]
 \centering
 \includegraphics[width=\linewidth]{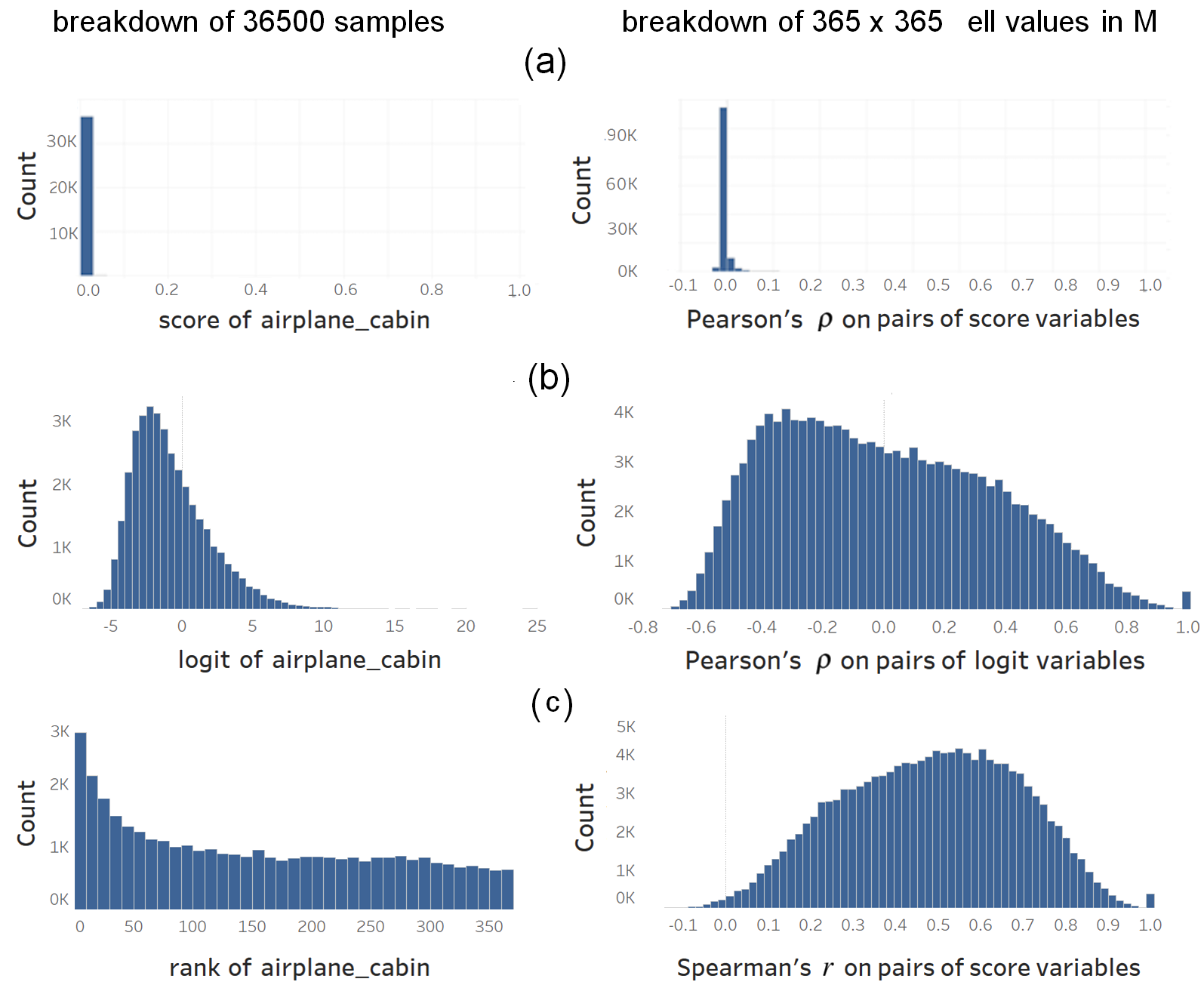}
 \caption {
The distribution of different types of input values (left) and of the corresponding score-based similarities (right) computed using: (a) Pearson's coefficient applied to probabilities, (b) Pearson's coefficient applied to logits, and (c), Spearman's rank coefficient applied to probabilities.
}
 \label{fig:statAnalysisBalanced}
\end{figure}

The output of classifiers is usually normalized into probabilities, with $1/ m$ being the average probability, where $m$ denotes the number of classes.
This makes the distribution of $P_j$ drastically skewed to the right.
Figure~\ref{fig:statAnalysisBalanced}a-left depicts this distribution for a random class in \texttt{Places365}.
Figure~\ref{fig:statAnalysisBalanced}a-right depicts the distribution if the similarity values in $M$, computed for all pairs of classes in the dataset.
This distribution drastically skewed to the right, which can mask global classification structure when visualizing $M$ as only small coherent clusters can emerge from the relatively few matrix cells corresponding to the long tail.
We examined two solutions to this problem:
\begin{itemize}
\item{\textbf{Using raw prediction scores (logits):}}
Figure~\ref{fig:statAnalysisBalanced}b-left depicts the distribution of these logits for a random class in \texttt{Places365}.
When used as random variables in Eq.~\ref{eq:sim-Pearson}, the logits result in a less skewed distribution of similarity values, compared with probabilities, as illustrated in Figure~\ref{fig:statAnalysisBalanced}b-right.
Logits are often used to circumvent similar problems~\cite{bucilua2006model, shrikumar2017learning}.
\item{\textbf{Using Spearman's rank coefficient:}}  This is equivalent to Pearson's coefficient applied to the rank values instead of the actual scores.
The distribution of the ranks over the range $[1, m]$ is more uniform (Figure~\ref{fig:statAnalysisBalanced}c-left) compared with probabilities, leading to near-normal distribution of the similarity values as illustrated in Figure~\ref{fig:statAnalysisBalanced}c-right.
\end{itemize}

The use of logits preserves fine-grained differences in the scores, whereas the use of ranks results in more rough similarity structures.
We recommend considering both options as we discuss in Section~\ref{sec:discussion}.

\begin{figure*}[!ht]
 \centering
 \includegraphics[width=0.98\linewidth]{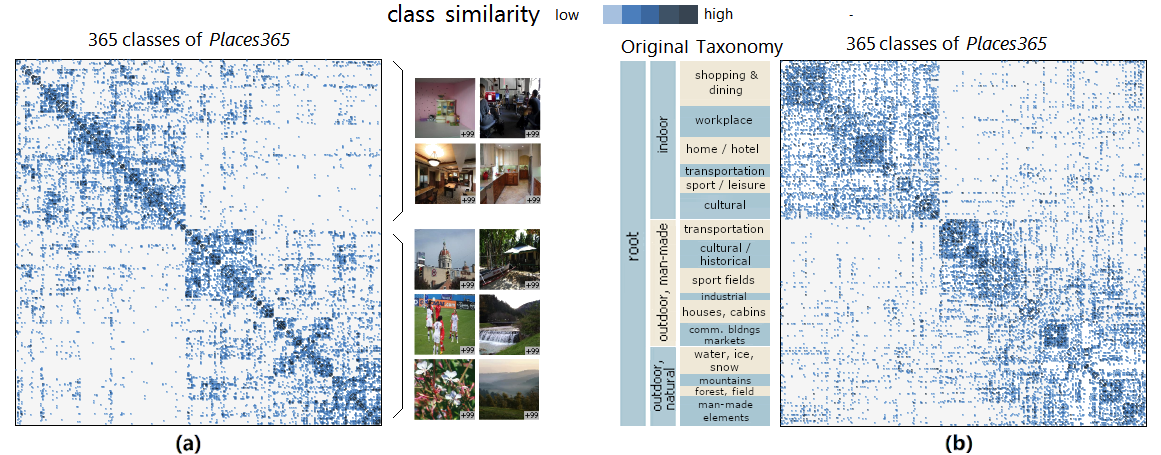}
 \caption {
		(a) The logit-based similarity matrix between the classes of \texttt{Places365}, ordered using \texttt{hclust}.
		The matrix exhibits two major groups of scene categories, indoor and outdoor environments.
		(b) The same matrix ordered according to the predefined taxonomy.
 }
 \label{fig:places365}
\end{figure*}

\section{Visualizing Class Similarity}
\label{sec:vis}
A 2D heatmap offers a natural way to visualize the similarity matrix $M$.
A key factor impacting the visualization is the ordering of the rows and columns.
Behrisch et al. surveyed seriation algorithms \cite{behrisch2016matrix}  illustrating the patterns they can expose in large matrices.
Our goal is to find a hierarchical structure over the classes.
Among available algorithms in $R$, we found that \texttt{hclust} can consistently reveal block patterns in various datasets if the values in the matrix follow a nearly-normal distribution.
We use complete linkage, the default agglomeration method in \texttt{hclust}.

Figure~\ref{fig:places365}a depicts the logit-based similarity matrix between 365 classes of  \texttt{Places365} described in Section~\ref{sec:statAnalysis}.
The \texttt{hclust} algorithm succeeds in revealing two major blocks along the diagonal, corresponding to rooms and outdoor scenes respectively.
In fact, the curator of the dataset selected their scene categories from a two-level scene taxonomy, depicted in Figure~\ref{fig:places365}b.
Parallel to our findings, this taxonomy divides the scenes into \emph{indoor} and \emph{outdoor} groups, with the later group further divided into \emph{man-made} and \emph{natural} outdoor scenes.
Ordering the matrix according to this taxonomy results in dense diagonal sub-blocks that correspond to its second level.
These sub-blocks largely correspond to the fine-grained ones found by  \texttt{hclust}.

\subsection{Recovering Hierarchical  Relations Among Classes}
\label{sec:recover_hierarchy}

\begin{figure}[ht!]
\includegraphics[width=\linewidth]{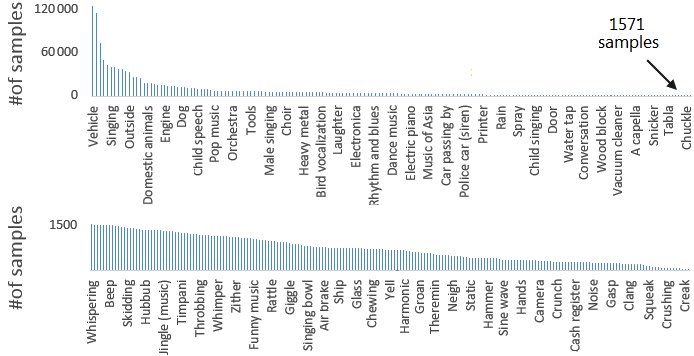}
\caption{
	The frequency of 527 classes in \texttt{AudioSet}, excluding \emph{Music} and \emph{Speech}, each appearing as labels in nearly 1 million samples.
	For better readability we print textual labels for a few classes only and split the long tail of the distribution.
 }
\label{fig:audioset_distrib}
\end{figure}

In multi-label datasets, the classes can vary in their level of abstraction, often leading to overlapping semantics.
This aims to maximize the prediction usefulness: The classifier can still predict correct high-level classes in case it fails to predict fine-grained ones.
An example of such dataset is \texttt{AudioSet} \cite{gemmeke2017audio}.
This dataset contains about 2 million audio clips, each assigned one or more labels from 527 categories that follow a predefined sound taxonomy.
Among the classes, 408 represent fine-grained categories such as \emph{Steel guitar}, while 119 are at multiple levels of abstraction such as \emph{Guitar}, \emph{Musical Instrument}, and \emph{Music}.
\texttt{AudioSet} has  a high degree of class imbalance as we demonstrate in Figure~\ref{fig:audioset_distrib}.

\begin{figure*}[!ht]
 \centering
 \includegraphics[width=\linewidth]{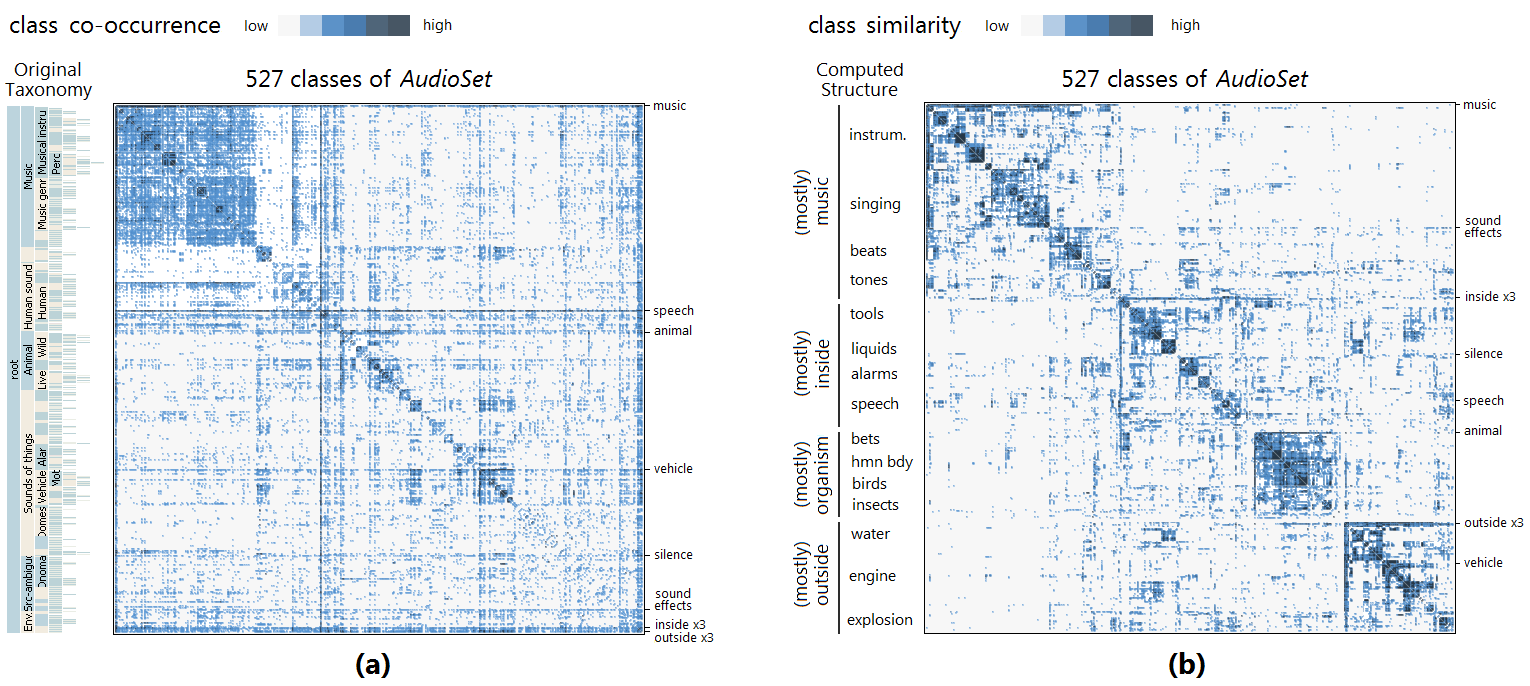}
  \caption{
	The classification structure in \texttt{AudioSet} based on
	(a) the co-occurrence matrix of the multi-labeled ground truth, ordered according to a predefined sound taxonomy, and
	(b) the logit-based similarity matrix, ordered using \texttt{hclust}.
	It reveals more coherent groups among the classes than the ones in (a) and surfaces high-level classes such as \emph{Inside} and \emph{Outside} that encompass other classes.
	}
\label{fig:teaser}
\end{figure*}

Figure~\ref{fig:teaser}a depicts the predefined taxonomy of the classes, along with their co-occurrence matrix in the multi-labeled ground truth.
The matrix suggests high co-occurrence within the \emph{Music} group, with other broad taxonomy groups being far less coherent.
The matrix further exhibits star patterns~\cite{behrisch2016matrix} in form of salient line crossings.
We mark the corresponding classes with a tick mark on the right side of the matrix.
The lines indicate that these classes exhibit high co-occurrence with a variety of other classes.
The lines stand out most for \emph{Music} and \emph{Speech}, each appearing in about $49\%$ of the samples.
The lines are also visible for \emph{Vehicle}, and \emph{Animal}, each also having multiple subclasses in the taxonomy.
Finally, a few subclasses of \emph{Inside} and \emph{Outside} form a dense band, visible in the bottom or right side of the matrix.
These subclasses serve as meta descriptors of sound events, leading to frequent co-occurrence with other classes.
The matrix falls short of exposing further structures due to its high susceptibility to class imbalance.

Figure~\ref{fig:teaser}b depicts the logit-based similarity matrix of the same dataset, with scores computed using VGGish \cite{hershey2017cnn}.
This matrix reveals several coherent and semantically-related groups among the classes that were masked in the co-occurrence matrix.
Furthermore, the matrix reveals \emph{flame} patterns \cite{lekschas2018hipiler} around major blocks along the diagonal.
These flames correspond to star patterns in the co-occurrence matrix, however, defining more accurate and informative spans over the classes.
We annotate the four major spans along the left side of the matrix.
The \emph{Music} and \emph{Animal} classes span two blocks that contains most of their subclasses in the original taxonomy, in addition to other semantically related classes that were under \emph{Sound of things} and \emph{Human}.
Interestingly, two new major blocks emerge that were marginalized in the original taxonomy, \emph{Inside} and \emph{Outside}.
The corresponding classes are grouped under  \emph{Acoustic environment} in the original taxonomy.
This suggests that whether the sound event takes place in indoor or outdoor environments plays a major role in the predictions made by the classifier.

\subsection{Recovering Split and Failed Similarity Groups}
\label{sec:splitgroups}

\begin{figure*}[!ht]
 \centering
 \includegraphics[width=\linewidth]{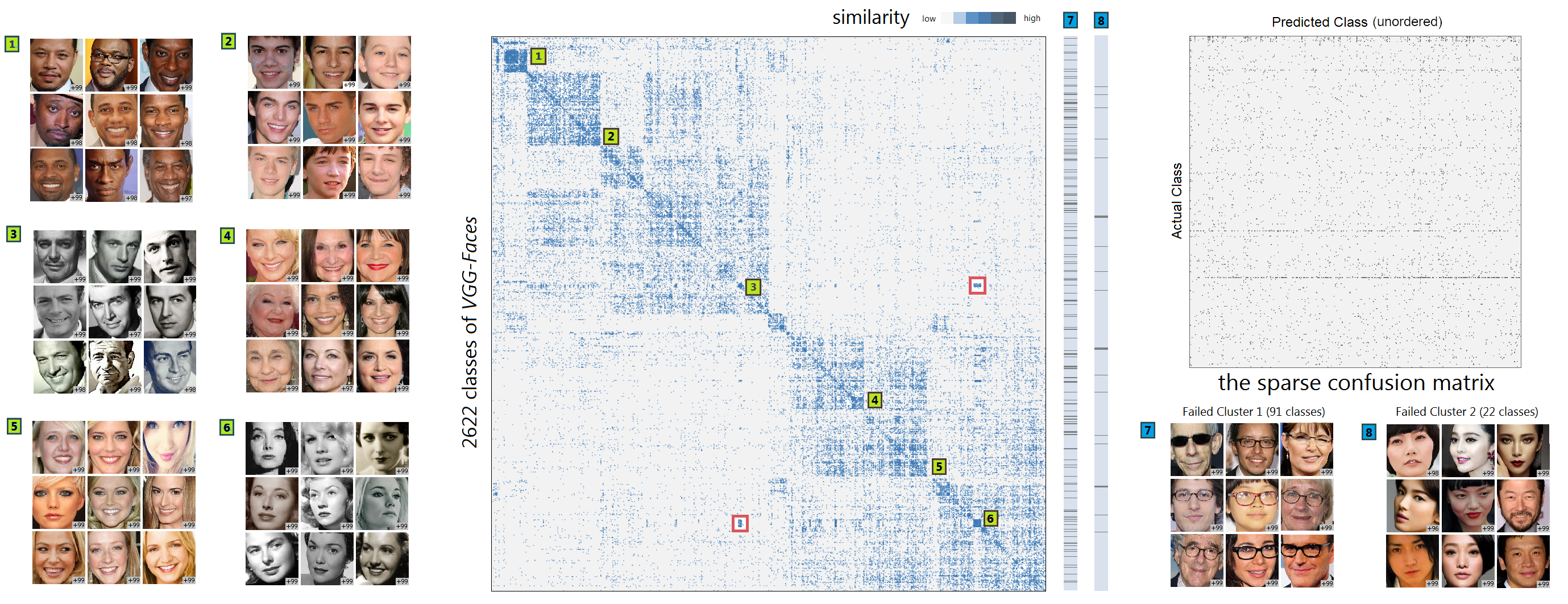}
 \caption {
	The logit-based similarity matrix between the classes of \texttt{VGGFace}, which contains images of 2622 celebrities.
	We annotate selected fine-grained clusters in green and show examples of the corresponding classes.
	These clusters do not surface in the confusion matrix.
	Two failed clusters are annotated in blue, along with corresponding columns that show their dispersion in the matrix.
 }
 \label{fig:vggfaces}
\end{figure*}

In many classification problems, overlapping class similarity groups might exist that do not fit in a strict hierarchy.
This happens, for example, with the \texttt{VGGFace} dataset whose classes represent 2,622 celebrities \cite{Parkhi15}.
Figure~\ref{fig:vggfaces} shows the logit-based similarity matrix computed over a balanced subset of 100 images per celebrity using the reference classifier provided by the dataset curators.
Ordering the matrix using \texttt{hclust} divides the classes into two broad similarity groups: male celebrities and female celebrities.
Within each group, the algorithm finds subgroups based on skin color, hair color, medium color (monochrome vs. multichrome), wrinkles, and other facial features.
These data features potentially define coherent similarity groups.
However, these groups are split as \texttt{hclust} identifies gender as the top level of the hierarchy.
In fact, the classifier rarely confuses celebrities of different gender for each other.
By looking at cases with such confusion, we found it often occurs with images scrapped for certain celebrities that did contain their opposite-sex spouses and went unnoticed when curating \texttt{VGGFace}.

Two dense off-diagonal clusters in the matrix are annotated with red boxes in Figure~\ref{fig:vggfaces}.
These clusters indicate high similarity between two subgroups of different gender that correspond to actors whose images are predominantly monochrome.
We refer to such related subsets as a \emph{split group}.
Dense off-diagonal sub-clusters are useful to identify split groups, however, they might not always emerge.
For example classes that represent celebrities who usually wear glasses turn out to have high similarity to each other.
Nevertheless, \texttt{hclust} disperses these classes across different subgroups in the matrix, as illustrated in an auxiliary column in Figure~\ref{fig:vggfaces}.
We refer to such groups as \emph{failed groups}.

Failing to consolidate potentially coherent groups is an inherent issue in hierarchical clustering, as it only allows inclusion and exclusion relations between the groups.
One solution to recover failed groups is to apply fuzzy clustering on the classes, with the rows of the similarity matrix $M$ serving as feature vectors.
This clustering paradigm can extract overlapping groups over the classes.
Figure~\ref{fig:vggfaces} depicts two recovered groups in \texttt{VGGFace} annotated as (7) and (8): glass wearers and celebrities of Asian ethnicity.
These groups suggest that the classifier develops corresponding facial features and relies on them to recognize the celebrities.
Such insights are very useful to infer potential corner cases of the classification model as we discuss in Section~\ref{sec:applicability}.

\section{Comparison with Related Work}
\label{sec:discussion}
{Class similarity measures} have been proposed for various purposes.
They have traditionally been based on raw input features \cite{roth2003going}, derived features \cite{opelt2006incremental}, or multi-labeled ground truth \cite{cisse2016adios, wu2015multi}.
Recent work has utilized learned features with cosine similarity~\cite{ye2017zero, hohman2019s}.
Nevertheless, little attention has been paid to visualizing the similarity matrix and analyzing the classification structure.
Likewise, prediction scores have been utilized to analyze classifiers, focusing however on their distribution~\cite{katehara2017prediction}, or calibration~\cite{guo2017calibration}.
The advantage of scores is that they reflect what the \emph{classifier} deems similar, offering a window to analyze its behaviour.
2D projections of the data space have been utilized to assess class separability~\cite{iwata2005parametric} and to reveal structures such as ones dictated by visual \cite{nguyen2016multifaceted} or linguistic features \cite{reif2019visualizing}.
Unlike matrices, scatter plots fall short of providing a scalable overview of hierarchical structures.

{Matrices} have been extensively used to analyze correlations \cite{bautista2016cliquecnn} or distances \cite{brasselet2009optimal} at the \emph{sample} level, or at the layer level \cite{gigante2019visualizing, li2016convergent, raghu2017svcca}.
We did not find prior work on utilizing matrices for class similarity, other than confusion matrices.
As discussed in Section~\ref{sec:rethinking}, our measure is more generic than confusion matrices:
In all of our examples, confusion matrices are either undefined (Section~\ref{sec:recover_hierarchy}), sparse (Figure~\ref{fig:vggfaces}-right), or fail to reveal coarse groups as we demonstrate in the provided notebook.

\section{Applicability and Future Work}
\label{sec:applicability}

Understanding structure can inform additional supervision to regularize neural classifiers~\cite{alsallakh2018Understanding}.
An earlier work demonstrates how a confusion matrix enables spotting subtle data-quality issues, as they pop out as outliers off the diagonal blocks \cite{alsallakh2017convolutional}.
This type of analysis enabled us to find various quality issues in \texttt{VGGFace} such as duplicate identities, hardly distinguishable twins, and mistaking celebrities for their spouses.
Furthermore, based on the similarity groups we found in that dataset (Figure~\ref{fig:vggfaces}), we identified potential dependence of certain classes on specific image features such as monochrome input or eye glasses.
We curated external images of the corresponding celebrities that did not have the respective features, for example, face images without eye glasses of celebrities in group (7).
The prediction scores for these images dropped significantly, leading to frequent misclassification.
Finally, the  structure enables comparing the established groups w.r.t. robustness to perturbations or to adversarial attacks.


Matrices can reveal coarse similarity groups between thousands of classes, especially when equipped with cut-off sliders and other visual boosting techniques~\cite{oelke2011visual}.
On the other hand, computing and reordering large matrices can be resource intensive, however, within feasible limits: about 2GB of memory and a few minutes of compute on an average CPU suffice for the examples we presented.
Nevertheless, matrix seriation remains an open challenge as multiple orderings are plausible.
While we recommend \texttt{hclust} for an initial overview, we encourage exploring different linkage methods, ordering algorithms, and correlation coefficients, examining the distribution of computed similarities, and employing further algorithms to extract potentially failed groups.

Our future work aims to provide interactive means to ease examining alternative ordering and failed groups.
We also aim to apply our approach to extreme and few-shot classification in order to analyze the behavior of classifiers designed to address long-tail distributions \cite{chen2019closer, wang2017learning}.
Thanks to its ability to handle class imbalance, our approach can help in analyzing how classes with few shots are affiliated with more common classes. 
Finally, our approach can help understand the evolution of classification structure over multiple layers in neural classifiers.

\section{Conclusion}
\label{sec:conclusion}

We presented means to compute and analyze class similarities in large-scale  classification based  on prediction scores.
Our means can handle datasets involving class imbalance, multi-labeled samples, as well as datasets with few prediction errors or scarce labels.
We provided extensive means to visualize the class similarity matrix and illustrated with four datasets how it can expose various types of relationships between the classes.
Our analytical and visual means inform the analysis of class similarities in a way that has not been formally addressed in the literature.

\bibliography{main}
\bibliographystyle{icml2020}

    \onecolumn

\section*{Appendix}
Here we present two further examples on visualizing classification structure.
\begin{figure}[hb!]
    \centering
    \includegraphics[width=\linewidth]{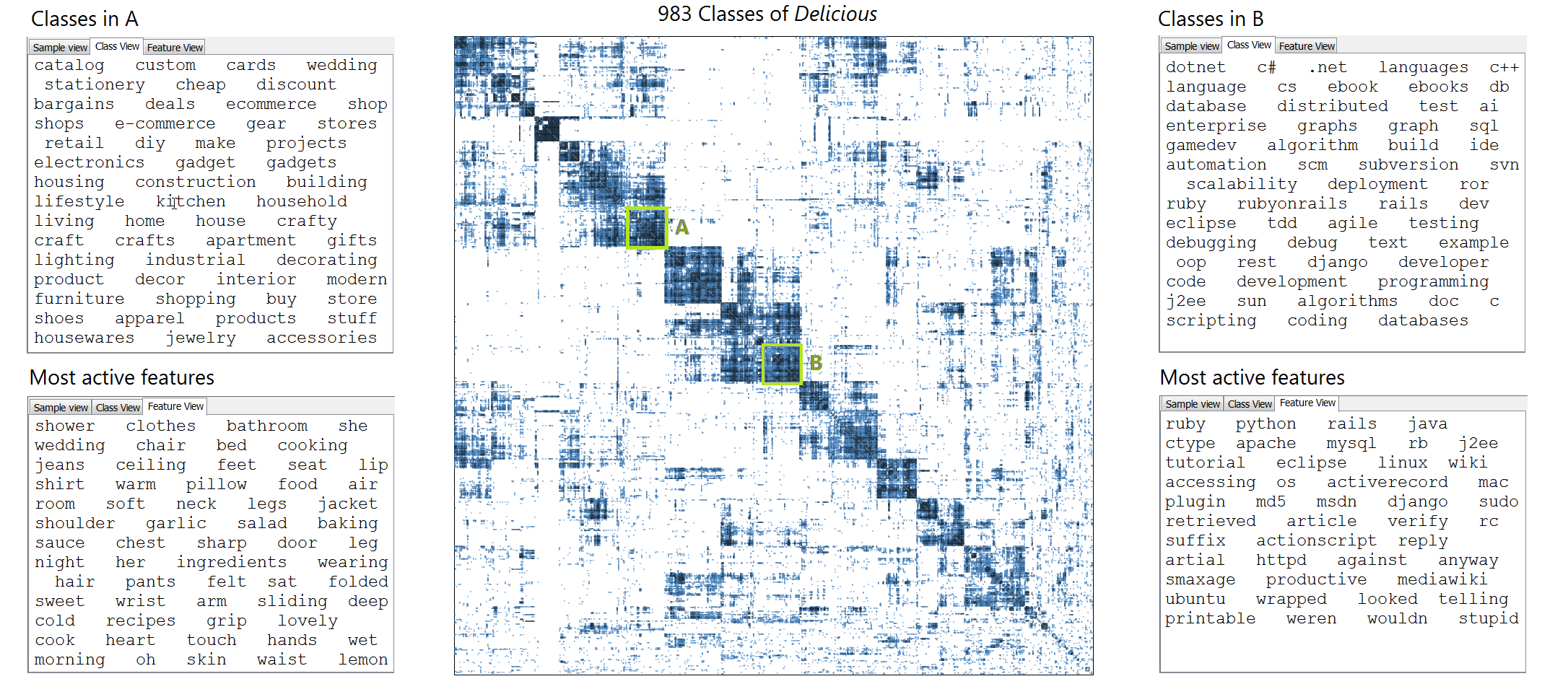}
    \caption{The probability-based similarity matrix of the \emph{Delicious} text classification dataset~\cite{tsoumakas2008effective}.
    The scores are computed using FastXML \cite{prabhu2014fastxml} and the similarities are computed using Spearman's rank coefficient.
    Two groups of classes are highlighted whose labels are listed in the respective text boxes.
    Group A contains predominantly shopping-related tags, while group B captures programming-related tags.}
\end{figure}
\begin{figure}[ht!]
    \centering
    \includegraphics[width=0.6\linewidth]{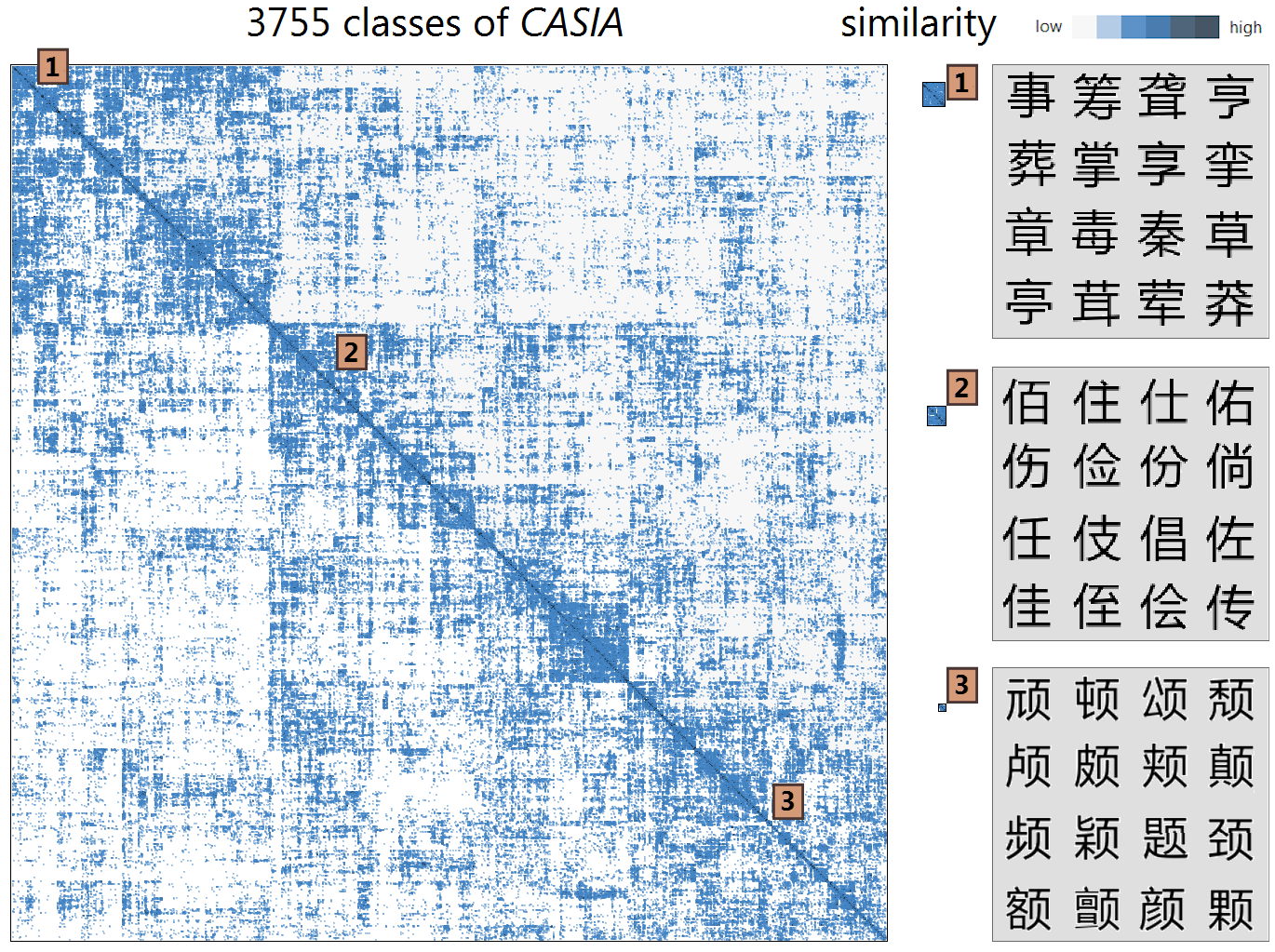}
    \caption{logit-based similarity matrix between 3,755 Chinese characters from the CASIA validation dataset \cite{liu2011casia}. The dataset contains 60 samples per Chinese characters which we classify using  Chinese offline handwriting classifier~\cite{CASIA_CNN} trained on the CASIA training set.
	The matrix reveals a few coarse groups, each containing a variety of coherent fine-grained groups along the diagonal, three of which are annotated  along with examples of their classes (depicted with a typeface).
	The character classes in the fine-grained groups share highly similar or identical strokes, e.g., a horizontal line in the top (1) or a shared stroke on the left (2) or right side (3).}
\label{fig:casia}
\end{figure}

\end{document}